\RequirePackage{fix-cm}
\documentclass[10pt,letterpaper,twocolumn]{article}
\usepackage[T1]{fontenc}
\usepackage{microtype,graphicx}
\usepackage[table]{xcolor}
\usepackage{amsmath,amssymb}
\usepackage{booktabs,array,multirow}
\usepackage{tikz,pgfplots}
\usepackage{algorithm,algorithmic}
\usepackage{caption,placeins,textcomp,needspace,balance}
\usepackage{research_layout}
\usepackage[hyperfootnotes=false]{hyperref}
\usepackage{xurl}
\usetikzlibrary{arrows.meta,positioning,calc}
\pgfplotsset{compat=1.16}
\definecolor{plotblue}{HTML}{326B8F}
\definecolor{plotorange}{HTML}{BF7330}
\definecolor{plotteal}{HTML}{337D77}
\definecolor{plotpurple}{HTML}{79589A}
\definecolor{plotred}{HTML}{AA5148}
\pgfplotsset{paperaxis/.style={
  tick label style={font=\scriptsize},label style={font=\small},
  legend style={font=\scriptsize,draw=none,fill=none,inner sep=1pt},
  axis line style={black!55},tick style={black!55},
  axis x line*=bottom,axis y line*=left,
  ymajorgrids,grid style={black!10},
  scaled ticks=false,/pgf/number format/1000 sep={\,},
  every axis plot/.append style={line width=0.9pt,mark size=2pt},
  clip=false}}

\newcommand{\BaselinePositions}{30{,}720}
\newcommand{\AggregateRatio}{10.67}
\newcommand{\NativePositions}{262{,}144}
\newcommand{\AggregatePositions}{327{,}680}

\newcommand{\NativeTG}{5.986}
\newcommand{\AggregateTG}{9.735}
\newcommand{\FinalTG}{18.284}
\newcommand{\FinalPeak}{15{,}626}

\newcommand{\NativeTGSpread}{0.86}

\newcommand{\sys}{JustFit}
\newcommand{\kvexec}{KVExec}
\newcommand{\swap}{PhaseSwap}
\newcommand{\statetrans}{StateTrans}
\newcommand{\code}[1]{\texttt{#1}}
\makeatletter\let\inputrows\@@input\makeatother

\newcolumntype{L}[1]{>{\raggedright\arraybackslash}p{#1}}

\hypersetup{hidelinks,pdftitle={JustFit: Just-in-Time State Management for Local LLM Serving},pdfauthor={Yuhua Chen},pdfsubject={KV execution, component residency, and state-preserving serving transitions}}
\renewcommand{\topfraction}{.88}\renewcommand{\textfraction}{.12}
\begin{document}
\twocolumn[\thispagestyle{plain}
\begin{center}
\hrule height1pt\vspace{12pt}
{\Large\bfseries JustFit: Just-in-Time State Management for Local LLM Serving\par}
\vspace{12pt}\hrule height1pt\vspace{17pt}
{\bfseries Yuhua Chen\par}\vspace{3pt}
{\small\href{https://www.yuhuachen.com}{www.yuhuachen.com}\par}
{\small\href{https://github.com/YuhuaBillChen/justfit-mlx/tree/justfit-repro-v4/examples/justfit}{Code, Quick Start, and results: JustFit reproduction kit v4}\par}
\vspace{17pt}
\end{center}]
\raggedbottom
\begin{abstract}
Local agents need memory for model execution and working history. We present \sys{}, an MLX runtime that coordinates their overlapping allocations: \kvexec{} executes and checkpoints four-bit KV with bounded workspace, \swap{} loads phase-dependent components, and \statetrans{} preserves history across execution modes. With a 27B Qwen3.8 MXFP4 model on a 24 GiB M4 Pro laptop, \sys{} reaches a 320K (\AggregatePositions{}-slot) occupied page-capacity high-water across two requests, \AggregateRatio{}$\times$ the \BaselinePositions{}-position single-request record of the evaluated historical mlx-vlm configuration. One request completes the full \NativePositions{}-position native window at median \NativeTG{} tokens/s. Each shape completes 16,384 outputs per request in three fresh processes: B1 uses one cold build and two prefix extensions; B2 uses three ordered prefix extensions (Section~\ref{sec:setup}). Image encoding can proceed while preserving a live 196,608-input text request. A controlled, repetitive 32K+6K workload reaches median \FinalTG{} tokens/s at \FinalPeak{} MiB; a separate AIME 2026 evaluation scores 29/30. Coordinating execution and state lifetimes makes longer histories feasible on personal hardware.
\end{abstract}

\section{Introduction}
Capable open-weight models make local coding and reasoning worth pursuing. Qwen3.8-27B reports 89.2 on GPQA Diamond and 90.3 on LiveCodeBench v6, compared with 91.3 and 88.8 for Opus 4.6 Max in the same upstream evaluation \citep{qwen38card}. Running such a model on a personal laptop would keep code, documents, and intermediate work on the user's machine. The challenge is to retain working context and sustain execution, not merely load the weights.

\begin{figure}[t]
\centering
\begin{tikzpicture}
\begin{axis}[paperaxis,width=.96\columnwidth,height=1.85in,
 ybar stacked,bar width=29pt,xmin=-.55,xmax=2.55,ymin=0,ymax=400,
 xtick={0,1,2},xticklabels={{mlx-vlm\\B1 baseline},{JustFit\\B1},{JustFit\\B2 aggregate}},
 xticklabel style={font=\fontsize{8}{9}\selectfont,align=center},
 ytick={0,64,128,192,256,320},ylabel={Completed / occupied slots (K)},
 legend style={at={(0.5,1.04)},anchor=south,legend columns=2,column sep=5pt}]
\addplot+[fill=plotblue,draw=plotblue,mark=none] table[x=x,y=input_k,col sep=comma]{data/hero_capacity_v25.csv};
\addplot+[fill=plotteal!50,draw=plotteal!80!black,mark=none] table[x=x,y=output_k,col sep=comma]{data/hero_capacity_v25.csv};
\legend{Input,Generated output}
\node[font=\small,anchor=south] at (axis cs:0,34) {30K};
\node[font=\fontsize{8}{9}\selectfont,anchor=south,align=center] at (axis cs:1,260) {256K full window\\\NativeTG{} tok/s};
\node[font=\fontsize{8}{9}\selectfont,anchor=south,align=center] at (axis cs:2,324) {320K page slots\\\AggregateTG{} tok/s total};
\end{axis}
\end{tikzpicture}
\par\vspace{8pt}
\pgfplotstableread[col sep=comma]{data/memory_growth_slopes.csv}\growthdata
\pgfplotstablegetelem{0}{slope_mib_per_1k}\of\growthdata
\let\baselineslope\pgfplotsretval
\pgfplotstablegetelem{1}{slope_mib_per_1k}\of\growthdata
\let\justfitslope\pgfplotsretval
\begin{tikzpicture}
\begin{axis}[paperaxis,width=.89\columnwidth,height=1.14in,
 xbar,bar width=11pt,xmin=0,xmax=205,ymin=-.55,ymax=1.55,y dir=reverse,
 xtick={0,50,100,150,200},ytick={0,1},
 yticklabels={{mlx-vlm\\8--24K},{Earlier JustFit\\8--64K}},
 yticklabel style={font=\fontsize{8}{9}\selectfont,align=right},
 xlabel={Peak growth (MiB per 1K input positions)},
 xlabel style={font=\fontsize{8}{9}\selectfont},
 ymajorgrids=false,xmajorgrids=true,
 tick label style={font=\fontsize{8}{9}\selectfont}]
\addplot[fill=plotorange,draw=plotorange,bar shift=0pt] coordinates {(\baselineslope,0)};
\addplot[fill=plotblue,draw=plotblue,bar shift=0pt] coordinates {(\justfitslope,1)};
\node[anchor=west,font=\fontsize{8}{9}\selectfont,xshift=2pt] at (axis cs:\baselineslope,0)
 {\pgfmathprintnumber[fixed,precision=1]{\baselineslope}};
\node[anchor=west,font=\fontsize{8}{9}\selectfont,xshift=2pt] at (axis cs:\justfitslope,1)
 {\pgfmathprintnumber[fixed,precision=1]{\justfitslope}};
\end{axis}
\end{tikzpicture}
\caption{\textbf{More usable context on the same laptop.} Top: 256K B1 (8.53$\times$ the 30K single-request record of the evaluated historical mlx-vlm configuration) and 320K occupied page-capacity slots across two B2 requests (\AggregateRatio{}$\times$ that record numerically, not a like-for-like single-request gain). B1 uses one cold build plus two prefix extensions; B2 uses three ordered prefix extensions. All complete 16K output per request below the 21,000-MiB guard. Bottom: earlier 64-output probes estimate process-peak growth over different labeled ranges, not KV-array bytes or the final binary's memory slope. K $=1{,}024$.}
\label{fig:hero}
\end{figure}
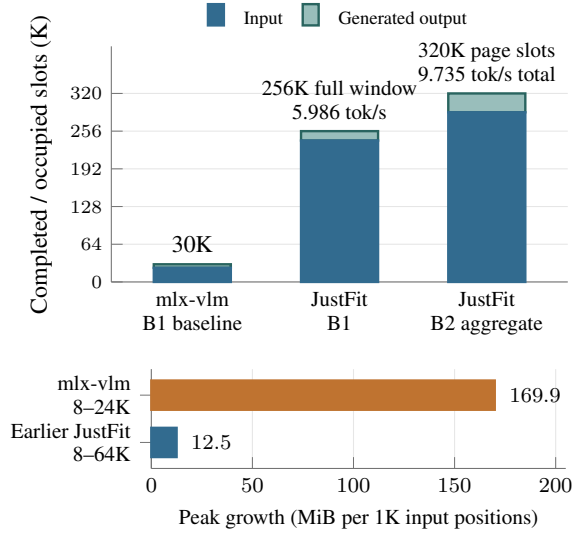

\paragraph{Context is an agent's working memory.}
Repository-level work spans source files, interfaces, test output, failed edits, and earlier decisions. Even selected files can exceed 100,000 tokens: in SWE-bench, Claude 2's 100,000-token window covers only 96.4\% of inputs built from the files edited by the reference fix \citep{swebench}. An agent also accumulates tool results as it works. When memory forces compaction, an extra model call summarizes the history; aggressive summaries can omit details needed later \citep{contextengineering,hermescompression}. Hermes already specifies at least 64,000 context tokens for tool-using sessions \citep{hermesproviders}. Larger retained context therefore gives long-running agents more room before capacity forces them to summarize or reread. Prefix caching complements it by reusing unchanged history rather than processing that history again \citep{sglang}.

\paragraph{Unified memory creates an opportunity and a constraint.}
Apple's M-series architecture lets CPU and GPU access the same memory, and MLX exposes the same arrays to both processors without explicit copies \citep{mlxmemory}. There is no separate GPU memory bank, but CPU-side staging, GPU workspaces, weights, and request state compete for one budget. Unified memory removes a placement barrier; it does not remove the need to control simultaneous allocations.

Four-bit (Q4) weights make this setting practical. MLX supports both affine integer quantization and MXFP4, a block-scaled floating-point format; scales and, for affine quantization, offsets add storage beyond the nominal four bits \citep{mlxquantize}. AWQ uses activation information to improve low-bit weight quantization \citep{awq}, while Qwen3.8-derived ternary Bonsai 2 27B pushes weight compression further \citep{bonsai2}. These are complementary to our contribution: we hold an MXFP4 checkpoint fixed and optimize the state around it. Its non-vision target tensors occupy 13.31 GiB, while an FP16 key/value (KV) cache would add 16 GiB at Qwen's native 256K window. Both cannot fit in 24 GiB before execution even begins.

\paragraph{Compression must survive execution.}
The KV cache retains the attention information for past tokens. Reducing it to four bits saves space, but a poorly chosen quantizer can perturb the attention values reused throughout a long generation. We use \emph{TQ4}, the runtime's four-bit cache based on TurboQuant's normalized, rotated mean-squared-error (MSE) formulation. TurboQuant reports full-precision-level retrieval in its long-context tests and strong low-bit generation results \citep{turboquant}; our integrated TQ4 runtime scores 29/30 on AIME 2026. This makes low-distortion compression a useful starting point. The remaining problem is execution: attention can reconstruct large floating-point operands, checkpointing can duplicate an entire compressed cache, and inactive components can remain resident while another phase needs their memory.

\sys{} manages these overlaps in time. \textbf{KVExec} keeps KV compact through attention and checkpointing using custom fused reconstruction and bounded temporary storage. \textbf{PhaseSwap} makes model components available for their required phases. \textbf{StateTrans} preserves request state as a single speculative decoder becomes a batch or temporarily makes room for image encoding. Together they reach 320K occupied page-capacity slots across two requests, while a single request reaches the model's full 256K native window (Figure~\ref{fig:hero}). The contribution is not another weight quantizer: it is making compressed state remain affordable throughout a working local service.

\section{Background: One Memory Budget, Different Lifetimes}
\label{sec:background}
\paragraph{State grows differently from weights.}
A request first processes its input (\emph{prefill}), then generates tokens (\emph{decode}). Its KV cache stores earlier keys and values so later attention does not recompute them. Qwen3.8-27B has 16 full-attention layers, each with four KV heads of dimension 256, interleaved with 48 Gated DeltaNet layers \citep{qwen38card}. Full-attention KV grows with context length; the recurrent layers instead maintain fixed-size state per active request. Model weights are shared among requests, but their histories are generally different.

\paragraph{Unified memory makes overlap the constraint.}
Because CPU and GPU access the same MLX arrays, moving an object from GPU work to CPU work does not create a second memory budget \citep{mlxmemory}. A conceptual accounting of simultaneous allocations is
\begin{equation}
 M(t)=W(t)+K(t)+S(t)+X(t)+H(t),
 \label{eq:memory}
\end{equation}
where $W$ is model-component storage, $K$ the compressed KV backing, $S$ recurrent and speculative state, $X$ temporary computation buffers, and $H$ other host and allocator storage. Aliased storage is counted once. Quantization reduces stored weights and KV; \sys{} additionally changes $W(t)$, $S(t)$, $X(t)$, and their overlap. The aim is to lower $\max_t M(t)$ without discarding useful history. Experiments measure macOS physical process footprint separately; the equation is not a measured attribution of that counter.

\paragraph{Compact histories need reusable storage.}
TQ4 stores each 256-element vector as 128 packed bytes plus a two-byte norm. Its payload per retained token position is
\begin{equation}
 \beta_{\mathrm{TQ4}}=16\cdot2\cdot4\cdot(128+2)
 =16{,}640\ \mathrm{B/position}.
 \label{eq:kvbytes}
\end{equation}
Thus 256K and 320K histories require 4,160 and 5,200 MiB of packed arrays. Including norms, TQ4 uses 25.4\% of the corresponding FP16 KV payload. To share that storage between growing requests, a \emph{paged KV pool} reserves backing arrays and assigns small blocks to each request, following the block-allocation idea of PagedAttention \citep{pagedattention}. A page table maps a request's token order to those blocks. JustFit uses 256-token pages; completion returns page IDs for another request to use rather than allocating a new pool.

\paragraph{Stored format is not execution format.}
Our prefill path uses MLX's optimized scaled dot-product attention (SDPA) with floating-point query, key, and value arrays. TQ4 must therefore be reconstructed for this consumer; our decode kernel instead reads compressed pages directly. Reconstructing $g$ KV heads at history length $T$, head dimension $d$, and $s$ bytes per scalar needs
\begin{equation}
 B_{\mathrm{group}}(T,g,s)=2gTds.
 \label{eq:layerbytes}
\end{equation}
At 240K input, all four BF16 heads need 960 MiB for K/V alone. Processing two heads at a time needs 480 MiB, leaving more room beside the same persistent history. This is the reason for grouped reconstruction, not a change to attention's context length.

MLX also delays computation: Python array operations first construct a computation graph. Calling \code{mx.eval} executes the pending operations needed to produce the specified arrays \citep{mlxlazy}. JustFit uses these boundaries to finish a group or layer before building the next, so old temporary dependencies do not accumulate. ``Evaluation'' below means executing this graph, not measuring model quality.

\section{JustFit Design}
\label{sec:design}
\begin{figure*}[t]
\centering
\resizebox{\textwidth}{!}{%
\begin{tikzpicture}[x=1cm,y=1cm,>=Latex,
 font=\sffamily\fontsize{9}{11}\selectfont,
 module/.style={draw,line width=.8pt,align=left,inner sep=7pt,text width=4.5cm,minimum height=1.65cm},
 store/.style={draw,align=center,inner sep=5pt,text width=4.5cm,minimum height=.65cm},
 edge/.style={->,line width=.85pt},
 note/.style={font=\sffamily\fontsize{8}{10}\selectfont,align=center,text=black!70,fill=white,inner sep=2pt}]
\node[module,draw=plotteal,fill=plotteal!5] (st) at (2.5,0)
 {\textcolor{plotteal}{\bfseries StateTrans}\\[3pt]
  Admit peers; choose MTP / AR\\Preserve live request state\\Reclaim after pending work};
\node[module,draw=plotorange,fill=plotorange!5] (ps) at (8,0)
 {\textcolor{plotorange}{\bfseries PhaseSwap}\\[3pt]
  Supply required components\\Keep active execution leases\\Release at safe boundaries};
\node[module,draw=plotblue,fill=plotblue!5] (kv) at (13.5,0)
 {\textcolor{plotblue}{\bfseries KVExec}\\[3pt]
  Decode from packed Q4 pages\\Reconstruct one head group\\Save and restore compact KV};
\draw[edge,plotorange] (st.east) -- (ps.west);
\draw[edge,plotblue] (st.north east) -- ++(0,.75) -| (kv.north);
\node[note] at (8,1.65) {Active rows and execution mode};
\node[note,above=2mm of st] {Arrivals, completion, cancellation};
\node[store,draw=plotorange!70,fill=plotorange!3,below=6mm of ps] (weights)
 {Head / input embedding\\MTP predictor / vision tower};
\node[store,draw=plotblue!70,fill=plotblue!3,below=6mm of kv] (pool)
 {Fixed Q4 backing pool\\per-request ownership};
\draw[edge,plotorange] (ps.south) -- (weights.north);
\draw[edge,plotblue] (kv.south) -- (pool.north);
\node[note,below=6mm of st,text width=4.5cm] {Required phase $\rightarrow$ components\\History survives phase changes};
\end{tikzpicture}}
\caption{\textbf{Three decisions under one memory budget.} StateTrans selects the active requests and execution mode; PhaseSwap supplies the components needed for that phase; KVExec reads and updates their compressed history. Request ownership persists while temporary components change. Figure~\ref{fig:kv-detail} expands the attention and checkpoint paths.}
\label{fig:system}
\end{figure*}
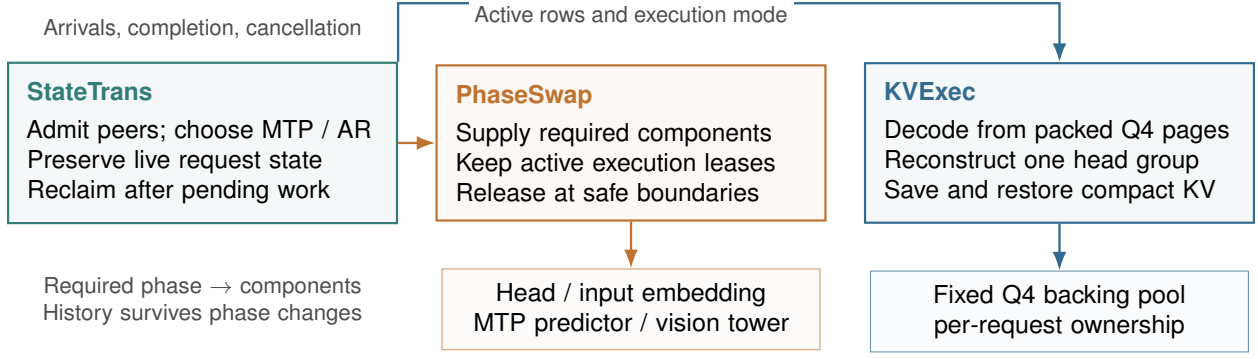

\sys{} separates the history a request must keep from the resources needed for its next computation. KVExec controls how compressed history is consumed, PhaseSwap controls which model components are present, and StateTrans decides when requests can share the device. Figure~\ref{fig:system} shows their interfaces.

The execution mode matters to all three. For one request, \emph{multi-token prediction} (MTP) uses a small predictor to propose tokens that the target model checks together. For several requests, \emph{autoregressive} (AR) batching already processes one next-token step for every active request in one forward pass. Our tested batched MTP is 1.7--1.9\% slower than AR at two requests and 12.1--20.4\% slower at three or four, with higher peaks (Appendix~\ref{app:batchedmtp}). JustFit therefore uses MTP for an eligible singleton and AR for two or more requests.

\subsection{KVExec: Executing Compressed KV}
\label{sec:kvexec}
TQ4 must preserve useful attention information without expensive calibration or per-token codebooks. Normalization separates a vector's magnitude from its direction; a randomized Hadamard transform (RHT) spreads large coordinates before nonuniform scalar quantization. For norm $\rho=\|x\|_2$, unnormalized Hadamard matrix $H_d$, and random-sign diagonal $S$, our TurboQuant-based encoding \citep{turboquant} is
\begin{equation}
 y=\frac{H_dS}{\sqrt d}\frac{x}{\rho},\qquad
 \widehat x=\widehat\rho\frac{SH_d}{\sqrt d}\,c[z].
 \label{eq:codec}
\end{equation}
The four-bit indices $z$ select 16 shared centroids; $\widehat\rho$ is the stored FP16 norm. The rotation and nonuniform levels reduce distortion at this small bit budget. KVExec contributes the execution paths around this existing representation: page access, fused reconstruction, and explicit control of temporary storage. The codec parameters are in Appendix~\ref{app:codec}.

\paragraph{Decode and prefill need different kernels.}
Decode usually supplies one new query per request. Expanding every cached token into a full floating-point array on each step would consume much of the saved memory, so our page-native attention kernel reads and reconstructs the Q4 data where it is used. Prefill supplies a block of queries, where MLX SDPA is a useful optimized consumer. For that path, \emph{we implement a custom Metal kernel} that combines page lookup, Q4 unpacking, centroid lookup, inverse rotation, and norm restoration. It writes only the final floating K/V operands, removing the contiguous packed-cache gather and the separate reconstruction arrays. Metal and MLX provide the programming and attention primitives; this fused page-to-operand path is part of JustFit.

\paragraph{Finish one head group before starting the next.}
To further reduce prefill headroom, the final path reconstructs two of the four KV heads at a time. Each group $j$ supplies its corresponding query heads:
\begin{equation}
 O_j=\operatorname{SDPA}(Q_j,\widehat K_j,\widehat V_j),\quad
 O=\operatorname{concat}_{\mathrm{heads}}(O_j).
 \label{eq:groupattention}
\end{equation}
Softmax operates independently per head, so the outputs concatenate without a cross-group attention reduction. We execute each group's pending work with \code{mx.eval} before creating the next group's operands, then execute the layer result before returning. The packed history remains available throughout. This halves the raw K/V operand at 240K input from 960 to 480 MiB. At the measured 192K mechanism probe, total incremental MLX peak including SDPA scratch falls from 3,123.8 to 1,589.3 MiB, with a 3.6\% latency increase (Section~\ref{sec:mechanisms}).

Speculative verification checks several proposed tokens at once. A short packed verifier is available for eligible two-to-four-query tiles; other eligible multi-query calls use grouped reconstruction and SDPA. The dispatcher gives direct inverse priority when both paths apply. Single-query attention thus avoids full-cache expansion; multi-query attention bounds workspace around SDPA. Appendix~\ref{app:codec} specifies the mask and shape conditions.

\subsection{PhaseSwap: Load Components for Their Next Use}
\label{sec:phaseswap}
The decoder backbone is needed throughout text generation, but not every component is. The output head turns the final hidden state into token scores; intermediate prefill does not need those scores. The vision tower is needed to turn an image into features, not to generate all subsequent text. PhaseSwap exploits these different lifetimes without moving the request's KV history.

The untied input embedding and output head each occupy 644.14 MiB of tensor payload---about 40,591 TQ4 positions in size. We detach the head during intermediate text prefill and restore it before first-token selection. The 215.21-MiB \emph{MTP predictor} is loaded only for single-request speculation and reuses the target embedding and head. This leaves temporary room without requiring smaller model weights.

\paragraph{Keep what another request still needs.}
Unloading after every locally idle step would cause repeated reloads during mixed prefill and decode. A \emph{lease} records an execution owner that still needs a component. If $O_c(t)$ is the owner set and $a_c(t)$ indicates that component $c$ is attached, the rule is
\begin{equation}
 |O_c(t)|>0\ \Longrightarrow\ a_c(t)=1.
 \label{eq:lease}
\end{equation}
Thus A's ongoing decode keeps the head available even during B's prefill chunks. A component becomes releasable after its last required use and owner have ended. StateTrans supplies the safe switching boundary. Here, resident means attached and materialized for use; detachment alone does not imply that its storage has been returned to macOS.

\paragraph{Encode images before growing their text history.}
An image request needs the BF16 vision tower (878.77 MiB of tensor payload) and its workspace only while extracting features. Keeping that tower through text prefill wastes headroom just as the new KV history grows. Waiting for an entire long text session to end also delays useful image work. Once admission permits it, JustFit encodes the image first, retains the resulting features, and releases the tower before processing image-text tokens.

This briefly pauses active text computation, but not its stored history. The worker finishes the current step, switches a single MTP request to batched AR, and releases the MTP predictor plus eligible head/embedding leases. Those temporary resources make room for the BF16 tower. After image features have actually been computed, the tower is released and text components are restored. Text decode and image-text prefill can then alternate on the GPU (Figure~\ref{fig:transitions}). The predictor stays detached while two or more requests are active, following the decode-mode policy above.

\paragraph{A detached field is not enough to free a weight.}
A cached callback can still hold the old embedding table after the model field has been cleared. We replace that strong table reference with a lookup through the model object. The old table can then be released, and subsequent input chunks use the restored table. This small ownership change makes the residency policy effective: the mixed-image trace records a 644.14-MiB drop in MLX-active allocation at embedding release. An image-feature cache hit skips tower loading altogether; cleanup restores the text components after an encoding error.

\subsection{StateTrans: Add Requests Without Rebuilding History}
\label{sec:statetrans}
A \emph{lane} is one active request slot. Continuous batching lets requests join and leave those slots while others keep generating. The scheduler must decide both whether another request fits and whether its preparation can run without a large interruption or allocation spike. StateTrans preserves target KV and recurrent state across these changes.

\paragraph{Switch execution mode without rebuilding the cache.}
A queued peer does not immediately stop MTP. Once the peer is selected for admission, the current speculative round finishes before the predictor is released. The original target KV and recurrent state continue in the AR batch rather than being copied or recomputed. When peers leave, the same state can return to MTP. This preserves the expensive history across changes in execution mode.

\paragraph{Let a new prompt progress without monopolizing decode.}
A long prefill can occupy both compute and workspace needed by ongoing generation. The released scheduler admits a fitting text request using lane availability and paged-KV reservations, without a separate length cap on its remaining input tokens. An earlier 8,192-token suffix gate could defer otherwise fitting prompts; it is disabled by default. Active decode leases keep the output head available, while chunking bounds the prefill work between decode opportunities. With active decoders, the worker runs four batched AR forwards, each advancing every active decode row, then permits at most 64 prefill tokens. We call this cadence \emph{N4/PF64}. Without a decoder, prefill chunks contain up to 256 tokens. Pending admitted prompts are ordered by ascending length, and a selected prefill runs to completion across its chunk opportunities rather than round-robining between prompts. This bounds work per opportunity, not total waiting time for an arbitrary workload.

\paragraph{Reserve output space as well as input space.}
Removing the suffix gate does not remove the output guarantee. With page size 256, prompt length $p_i$, requested output $o_i$, and guarantee $g_{\mathrm{out}}$, an unshared request reserves
\begin{equation}
 R_i=\left\lceil\frac{p_i+\min(o_i,g_{\mathrm{out}})}{256}\right\rceil
 \label{eq:reservation}
\end{equation}
page equivalents. The four-lane policy permits 16,384 output tokens while guaranteeing the first $g_{\mathrm{out}}=8{,}192$. Shared prefixes use reference counts. Later output consumes spare pages; if protected reservations are threatened, only requests already beyond their guarantee are eligible for explicit capacity termination, starting with the largest surplus. A fitting queued request may bypass a capacity-blocked one, but an aged blocked request can be overtaken at most eight times.

\paragraph{Give queued images an admission opportunity.}
An image waiting at least 500 ms becomes due for admission, provided a lane and KV reservation fit and the active retained-token sum passes the vision gate. The candidate gate is 198,656 actual input-plus-generated positions, not reserved output. At most one image enters per scheduler turn. If an overdue image cannot enter, later text arrivals do not bypass it again in that turn. Once eligible, it uses PhaseSwap's brief vision phase; afterward all active AR rows resume, with decode before the first image-text prefill chunk.

\begin{figure*}[t]
\centering\input{figures/state_transitions.tex}
\caption{\textbf{Requests keep their history while execution changes.} Text requests A--E share four lanes and 18 equal-capacity page groups; labels show illustrative group counts, not bytes or tokens. E reuses B's slots after pending work and the last reference end. After D leaves, eligible image F is encoded before its text prefill; A/C/E retain their KV and recurrent state through the text pause. Text components then return and all active AR rows advance at each decode step. A resumes MTP when peers finish and re-promotion gates pass. This event-order schematic omits decode growth, explicit reserved-page accounting, and elapsed time. It assumes admission gates pass; image measurements use one incumbent text lane (Section~\ref{sec:vision}).}
\label{fig:transitions}
\end{figure*}

\paragraph{Return storage only after its last use.}
Completion or cancellation first finishes the current step and pending cache writes, then removes the departing row and releases its page references. Zero-reference pages become reusable. Distinct owners for successive active request groups prevent delayed cleanup from releasing a newer group's components. When one eligible row remains, re-promotion is enabled, and no prompt is pending or newly admitted, MTP binds to the survivor's existing state. Algorithm~\ref{alg:transitions} summarizes this sequence.

\begin{algorithm}[t]
\caption{Changing the active requests safely}
\label{alg:transitions}
\small
\begin{algorithmic}[1]
\STATE Finish current step (complete MTP round if speculative)
\STATE Execute pending cache writes; remove departing rows
\STATE Release their references; recycle zero-reference pages
\STATE Select a peer $r$ that passes admission and fairness rules
\IF{$r$ exists}
  \IF{one active request uses MTP}
    \STATE Keep its target state; release predictor; switch to AR
  \ENDIF
  \IF{$r$ needs uncached image features}
    \STATE Pause text; release eligible head/embedding leases
    \STATE Load tower; compute image features; release tower
    \STATE Restore text embedding and output head
  \ENDIF
  \STATE Admit $r$ and continue the prefill/decode cadence
\ENDIF
\IF{one eligible row, re-promotion enabled, no pending/new prompt}
  \STATE Resume MTP on that row's existing target state
\ENDIF
\end{algorithmic}
\end{algorithm}

\subsection{Keep Checkpointing from Undoing the Savings}
\label{sec:persistence}
Automatic prefix caching (APC) saves reusable history for later calls. A bounded writer alone is insufficient if snapshot creation first copies the entire cache: the earlier path created a contiguous packed row before serialization began.

Our synchronous disk path instead reads \emph{views} of the existing page runs while the request retains ownership. Saving finishes before that request can mutate or release its pages. A single-pass safetensors writer streams bounded pieces to a checkpoint and atomically publishes it, avoiding a second temporary-file rewrite. Restoration writes packed runs into the existing pool. At 224K, borrowing rather than copying lowers incremental snapshot allocation from 3,640 to 227.5 MiB; the separate writer improvement roughly halves write bytes and save time (Section~\ref{sec:mechanisms}). Asynchronous callers still take an owning snapshot because their pages may change before writing finishes. The principle is the same as in attention: consume the compact state without first creating a whole-cache duplicate.

\section{Experimental Setup}
\label{sec:setup}
\paragraph{Platform and comparison.}
Experiments use an M4 Pro MacBook with 24 GiB unified memory and fixed Qwen3.8-27B MXFP4 weights. The evaluated historical mlx-vlm configuration completes 24K input plus 6K output and reaches the process guard at 32K+6K. JustFit's single-request target is the checkpoint's full 256K native window (240K+16K), while the shared-capacity workload is (160K+16K)+(128K+16K), or 320K across two requests. We write B1/B2 for one/two active requests and use K $=1{,}024$. The reference for both ratios is that historical configuration's 30,720-position single-request record: full-window B1 is 8.53$\times$ that record and B2's aggregate occupied page-capacity high-water is 10.67$\times$ numerically. The latter compares different request shapes and is not a single-request context multiplier.

For the released reproduction setup, the Metal host control \code{iogpu.wired\_limit\_mb} is set to 22,016 MiB. This wired-memory ceiling is distinct from the 21,000-MiB sampled process-footprint guard used in the capacity experiments. The former is a host-level requirement of the released setup; the latter is a fail-closed experiment policy below it. This statement describes the public reproduction configuration and does not retroactively assign that host setting to historical cohorts whose manifests did not record it.

\paragraph{Baseline scope.}
The baseline denotes a historical mlx-vlm execution snapshot, not the latest upstream release or an ideal FP16 reference. Its preserved 32K guard-run log reports an MTP drafter and a requested four-bit TurboQuant configuration; the launch manifest disables paged KV, segmented batch KV, and fused TQ reconstruction. A configured quantization flag alone does not establish every cache object's effective storage type. We therefore report the measured process footprint without treating this baseline as a separately verified FP16-cache implementation. The 16-GiB FP16 calculation in the introduction is an independent payload reference.

\paragraph{Complete generation at each final shape.}
Each final B1/B2 workload runs in three fresh processes and generates all 16K output tokens per request. B1 includes one complete cold input build and two extensions from a 161,792-position checkpoint, each processing 83,968 additional input positions. B2 begins from exact prefixes: its first trial processes 10,240/1 new positions, and the next two process 1/1. The longer request is submitted first; the second follows 0.5 s later. A separate pair of 128K+16K requests builds both inputs cold. We aggregate completed-state and decode results by workload; input and end-to-end times retain their measured work amounts (Appendix~\ref{app:evo10}).

\paragraph{Workloads and measurements across development.}
Capacity inputs repeat a word and use greedy selection with end-of-sequence (EOS) suppression to reach the requested output length. MTP applies the same static bias to draft and target. The harness checks token counts, output agreement, a subsequent request, and page reuse. The fixed 32K+6K study repeats each successful configuration in three fresh processes, with MTP verification width three: two proposed tokens and a bonus position. APC is disabled for stages 1--7 and enabled with fresh directories and zero input-prefix hits for 8--10. Earlier short-output, concurrent, AIME, and transition experiments retain their measured versions. Image admission is a later serving extension, evaluated separately from this development series.

\paragraph{Mixed image workload.}
One image arrives about 0.75 s after a long text request's first output token. Text generates 512 outputs; the image request has 453 prompt positions and up to 64 outputs, and must finish before the text request. The source PNG is $1049\times1176$; preprocessing limits area to 524,288 pixels, edge to 2,048, and vision microbatch to one. Each trial uses a fresh process and frozen-prefix clone. Experimental active-context gates are 143,360 for the 96K/128K arms and 204,800 for the 160K/192K/196K arms. The 198,656-position candidate default was chosen after these tests (Appendix~\ref{app:vision}).

\paragraph{Counters and timing.}
A watchdog samples macOS physical process footprint every 0.25 s, stopping at an integer-MiB sample of at least 21,000; the earlier 96K/128K image trials use 20,700 MiB. MiB, GiB, and GB mean $2^{20}$, $2^{30}$, and $10^9$ bytes. Prefill rate (PP) divides new input tokens by prefill work time. Decode rate (TG) divides post-first-token outputs by the generation interval; for concurrent requests it sums outputs only during their common active interval. End-to-end wall rate includes preparation and non-overlap. Pool high-water occupancy is $256P_{\mathrm{layer}}/16$, where $P_{\mathrm{layer}}$ counts occupied pages across the 16 attention layers. This counter reports page-capacity equivalents at 256-position granularity, not a per-token census of initialized KV entries; output counts independently establish completion. Repeated TG uses medians; peak ranges preserve the observed extrema. Mechanism probes separately measure incremental MLX allocation and application-level write bytes.

\section{Evaluation}
\label{sec:results}
\subsection{The Full Model Window and a Larger Shared Pool}
\label{sec:capacity}
JustFit reaches \AggregatePositions{} occupied page-capacity slots across two active requests, \AggregateRatio{}$\times$ the \BaselinePositions{}-position single-request record of the evaluated historical mlx-vlm configuration. This is an aggregate pool-occupancy comparison, not a per-request context multiplier. A single request fills Qwen's entire \NativePositions{}-position native window, while the larger B2 pool establishes system capacity beyond that per-request model configuration. Table~\ref{tab:capacity} reports complete generation at both endpoints.

\begin{table}[t]
\centering\small
\caption{\textbf{Complete generation with large retained histories.} B1 fills the native model window; B2 tests shared state. Every request generates 16K outputs. TG is median singleton or aggregate common-active-interval rate; peak is the full observed process range. Section~\ref{sec:setup} defines cold-build and exact-prefix initialization.}
\label{tab:capacity}
\setlength{\tabcolsep}{4pt}
\begin{tabular}{lrrr}\toprule
Workload & Complete & TG (tok/s) & Peak (MiB)\\\midrule
\inputrows data/generated/capacity_summary_rows.tex
\bottomrule\end{tabular}
\end{table}

All three 256K trials generate 16,384 tokens, reach the full occupied window, match their output sequences, and pass subsequent-request and page-reuse checks. Median TG is \NativeTG{} tokens/s; the 5.936--5.987 range spans \NativeTGSpread{}\% of the median. Whole-run peaks differ by 93 MiB. Prefix initialization thus avoids redundant early work while still exercising complete generation at the native window.

All three 320K trials likewise generate the full output for both requests and reach the same pool high-water mark. Median aggregate TG is \AggregateTG{} tokens/s, with a 9.732--9.738 range. A separate fully cold pair of 128K+16K requests completes 294,912 aggregate positions at 10.341 tokens/s; its instantaneous page-capacity high-water reaches 294,400 slots because one request finishes earlier. These are occupied pages observed during generation, not merely configured allocation limits; the counter does not measure exact initialized-token census.

\subsection{How the Design Changes Memory and Throughput}
\label{sec:tradeoffs}
\begin{figure*}[t]
\centering\input{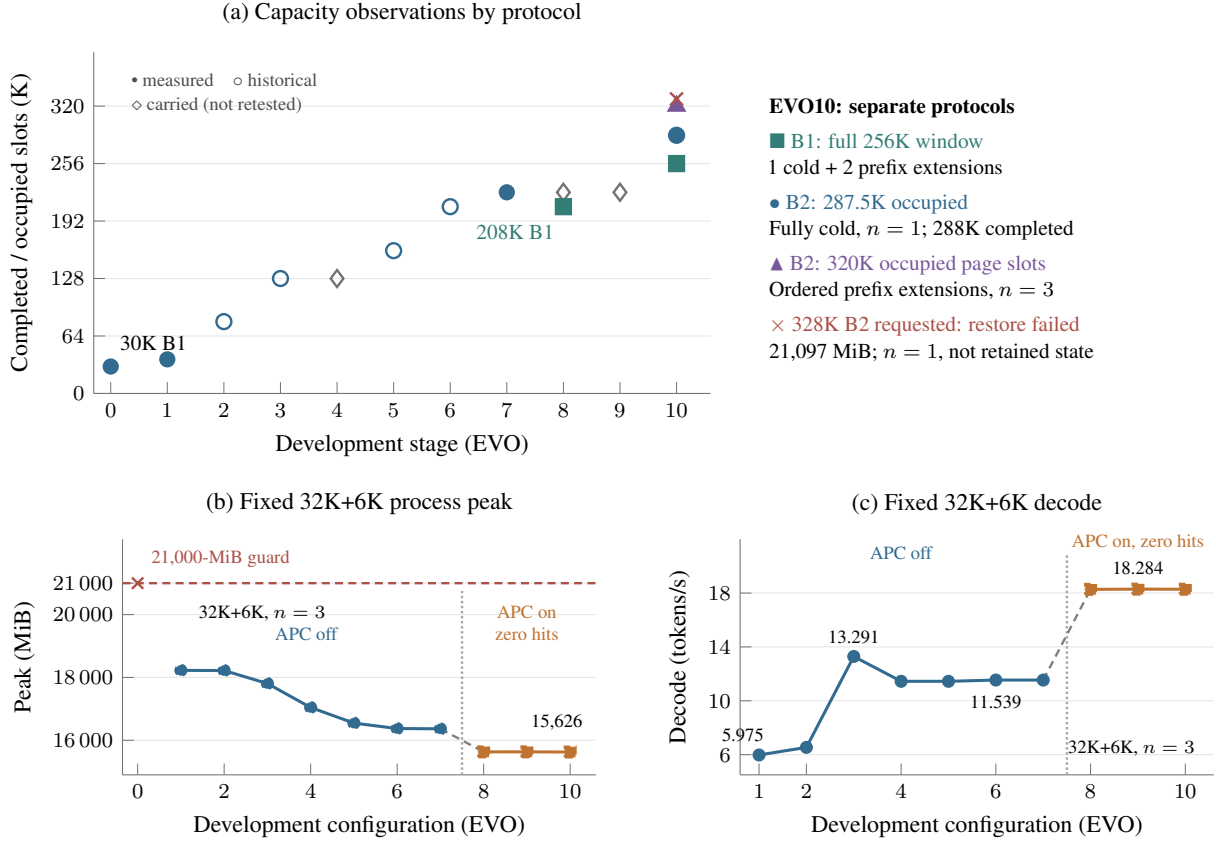}
\caption{\textbf{Capacity and execution across development.} (a) Separate markers distinguish measured, historical, and carried records; capacity was not retested at every stage. EVO10 shows B1's full 256K window, fully cold B2's 287.5K instantaneous occupancy (288K total completed work), and 320K ordered-prefix B2 occupancy. The red cross is a requested 328K shape that failed during restore, not retained capacity. Initialization and repeat counts accompany the markers. (b,c) Fixed 32K+6K medians and min--max from three processes per stage. The baseline cross marks guard termination. The dashed bridge marks APC changing from disabled (1--7) to enabled with zero hits (8--10); it is not an isolated verifier ablation. Table~\ref{tab:evo} defines the stages.}
\label{fig:evolution}
\end{figure*}

The fixed 32K+6K workload compares successive designs at the same input/output length (Figure~\ref{fig:evolution}). The baseline reaches the guard; the first completing TQ4+MTP configuration achieves 5.975 tokens/s at 18,225 MiB. The final system achieves \FinalTG{} tokens/s at \FinalPeak{} MiB: 3.06$\times$ faster decode and 2,599 MiB lower median peak than that first completing configuration. PP changes from 110.82 to 114.37 tokens/s.

The intermediate designs expose the tradeoff. Segmented KV reaches 13.291 tokens/s; the earlier paged path runs at 11.539 but uses less memory and supports shared state. The later configuration with the optimized verifier reaches 18.275; its APC protocol also changes, so the full difference cannot be attributed to the verifier alone. Head grouping and bounded checkpointing preserve approximately 18.28 tokens/s at 32K while enabling much larger histories. Their effect becomes important when a long attention reconstruction or whole-cache snapshot dominates the peak.

Figure~\ref{fig:hero}'s lower panel summarizes observed context-dependent peak growth using earlier 64-output probes. Fitting $F_{\mathrm{peak}}(L)=a+gL/1024$ gives 169.9 MiB per 1K input positions for the baseline over 8--24K, and 12.5 for JustFit over 8--64K. These whole-process slopes include workspace and allocation behavior. At the shared 8K+64 point, TG rises from 15.63 to 24.31 tokens/s with identical output tokens. Appendix~\ref{app:short} gives the probes.

\subsection{Reductions in Temporary Allocations}
\label{sec:mechanisms}
At 192K history, two-head reconstruction lowers incremental single-layer MLX peak from 3,123.8 to 1,589.3 MiB, including SDPA scratch. Latency changes from 271.220 to 280.897 ms, a 3.6\% increase; the selected two-head path matches the all-head reference bitwise at the tested lengths. 

At a 224K prefix, borrowing page-run views instead of extracting a contiguous copy lowers snapshot peak allocation from 3,640 to 227.5 MiB. A separate writer-only comparison lowers application-accounted writes from 7.63 to 3.82 GB and median save time from 2.976 to 1.507 s. Extraction and serialization must both be bounded: removing a second file write cannot remove a duplicate created before the writer is called. Appendix~\ref{app:mechanisms} reports the size sweeps.

\subsection{An Image Can Finish Before the Long Text Request}
\label{sec:vision}
The phase-separated path admits an image while a 196,608-input text request is already generating. The image completes in 21.636 s, before the text's 512-token output ends; both streams finish at a 20,852-MiB process peak below the 21,000-MiB guard. Table~\ref{tab:vision-main} reports tests from 96K to 192K text input.

\begin{table}[t]
\centering\small
\caption{\textbf{Image progress beside active text generation.} One completed trial per row; text output is 512 tokens and image prompt/output is 453/64. The image finishes first in every trial. Appendix~\ref{app:vision} supplies actual gate settings and the additional edge point.}
\label{tab:vision-main}
\setlength{\tabcolsep}{4pt}
\begin{tabular}{rrrr}\toprule
Text input & Image E2E (s) & Peak (MiB) & Guard (MiB)\\\midrule
\inputrows data/generated/vision_main_rows.tex
\bottomrule\end{tabular}
\end{table}

In the 192K trace, releasing the embedding lowers MLX-active allocation by 644.14 MiB. Tower load, embedding restore, and head restore take 0.151, 0.113, and 0.110 s in host calls. The maximum recorded text-stream gap is 1.177 s. The text request keeps its target pages through image encoding, shares AR execution with the image request, and then resumes MTP. The result demonstrates why request state and temporarily used model components need different lifetimes.

\subsection{Reusable Prefixes, Safe Departures, and Decode Choice}
\label{sec:reuse}
An earlier exact-prefix test restores 16,383 cached positions and processes one new position in an 82.314-ms restore/prefill interval while another request decodes. A mixed-arrival trace verifies MTP-to-AR-to-MTP on the same target state; three short requests finish before a blocked large request emits its first token. The first two short streams have P99 inter-token gaps of 89.8 and 89.2 ms. After a client disconnects at 128 stream events, surviving and new requests each complete 512 outputs and reuse its released pages.

The decode-policy comparison supports the implementation choice in Section~\ref{sec:design}. All six AR/batched-MTP pairs match outputs; high draft acceptance does not offset the tested batch-verification cost (Appendix~\ref{app:batchedmtp}). Short tests of the common component manager also preserve outputs and page reuse (Appendix~\ref{app:regression}).

\paragraph{Four active streams after admission repair.}
A later cold-input trial disables the legacy suffix gate, starts a 128K-input request, and submits three 8K-input peers after its first output. All four generate 12K tokens. With singleton MTP and concurrent AR, common-active aggregate TG is 22.570 tokens/s; including cold prefill, staggered admission, and the final survivor gives 12.147 output tokens/s over 4,046.521 s. Sampled peak is 18,915 MiB, leaving 2,085 MiB below the guard. Postflight and page reuse pass. This is one staggered-arrival qualification, not a matched speedup comparison against the earlier B4 runs (Appendix~\ref{app:b4current}).

\subsection{Extended Reasoning with Compressed KV}
\label{sec:quality}
In a separate AIME 2026 evaluation, paged TQ4 answers 29/30 problems correctly and generates 696,834 tokens at 15.04 token-weighted tokens/s, with a maximum recorded footprint of 19,043 MiB. Uniform INT8 answers 28/30 at 14.64 tokens/s. Both reach an output limit on one problem; their remaining difference is a sampled reasoning trajectory. This is direct end-to-end evidence for retaining useful reasoning with the chosen compressed representation. An earlier Bonsai 27B run with TQ4 also answers 26/30 on a MacBook Air. That measurement predates Bonsai 2. Appendix~\ref{app:quality} gives the evaluation protocol and error analysis.

\section{Related Work}
\paragraph{Model and cache compression.}
AWQ uses activation statistics to improve low-bit weight quantization, whereas MLX affine Q4 and MXFP4 specify stored values and scale metadata \citep{awq,mlxquantize}. Ternary Bonsai 2 27B illustrates further compression of the Qwen3.8 backbone \citep{bonsai2}. These advances reduce static model storage. KIVI, KVQuant, and TurboQuant instead compress the growing KV state \citep{kivi,kvquant,turboquant}. JustFit fixes both the target weights and a TurboQuant-based cache representation, then optimizes their execution and lifetimes.

\paragraph{Memory-constrained execution.}
FlexGen coordinates weights, activations, and KV across GPU, CPU, and disk \citep{flexgen}. LLM in a Flash, FlexInfer, PowerInfer-2, and ActiveFlow optimize limited-memory weight availability and I/O \citep{llmflash,flexinfer,powerinfer2,activeflow}. DwarfStar streams routed experts with bounded cache and prefill headroom \citep{dwarfstar}. Our dense-backbone setting instead shares one physical budget among persistent history and selected phase-dependent components. FlashAttention optimizes attention I/O \citep{flashattention}; Open-TQ-Metal uses compressed decode and floating SDPA prefill on Apple Silicon \citep{opentqmetal}. KVExec adds page-addressed fused reconstruction, sequential head-group execution, and compact persistence.

\paragraph{Serving and reusable history.}
Orca schedules iterations, PagedAttention allocates KV blocks, Sarathi-Serve interleaves chunked prefill with decode, and SGLang shares prefixes \citep{orca,pagedattention,sarathi,sglang}. Native Apple-silicon batching and persistent quantized agent memory are explored by vllm-mlx and Agent Memory Below the Prompt \citep{vllmmlx,agentmemory}. JustFit couples these serving concerns to component lifetimes, so mode changes and image admission preserve the request state already accumulated.

\section{Discussion and Limitations}
\label{sec:limits}
\paragraph{From capacity to useful interaction.}
B1 fills the native model window; beyond-native single-request operation requires a changed model configuration and separate evaluation. For a cached prefix $L$ and new suffix $\Delta$, a later turn requires
\begin{equation}
 T_{\mathrm{input}}=T_{\mathrm{lookup/restore}}(L)+T_{\mathrm{prefill}}(\Delta\mid L),
 \label{eq:incremental}
\end{equation}
not a full rebuild of $L+\Delta$. The 256K trials show close decode rates and peaks across cold construction and prefix extension. They do not repeat the same early input work; 320K peaks also range from 20,432 to 20,953 MiB across different initializations. 

\paragraph{Latency remains the practical constraint.}
A full 240K cold build takes 67.7 minutes, followed by 46.0 minutes for 16K generation. Reuse avoids repeated ingestion, but new tokens still attend to a long history. Apple reports 3.33--4.06$\times$ TTFT speedups for base M5 over base M4 at 4,096 input tokens \citep{applem5}; BaseRT targets M5 matrix-heavy prefill \citep{basertm5}. Such hardware progress and improved weight compression complement JustFit's scheduling of memory lifetimes. Performance on newer hardware needs direct measurement.

\paragraph{Quality and measurement scope.}
TurboQuant motivates a low-distortion representation; it is not lossless reconstruction of the original KV. Our AIME result checks extended reasoning, not full-window comprehension, repository-level success, or compaction frequency. Repetitive capacity inputs exercise allocation and complete generation with workload-specific MTP acceptance. Development stages combine mechanisms and an APC protocol change; isolated probes measure local allocations, not full-system attribution. Sampled physical footprint, MLX-active bytes, and host-call/I/O counters describe different quantities.

The largest B2 trial leaves 47 MiB of sampled margin; these single-platform runs and the 250-ms watchdog do not establish continuous-service reliability. The newer four-stream text trial has greater headroom but is a single workload, not a general latency or reliability guarantee. Image tests use one bounded image beside one text request, with 512/64 outputs; video, multiple images, and B4-plus-vision remain outside the measured scope. Image age and bounded bypass limit overtaking, not wall-clock waiting for unavailable capacity. The 200,704-input edge has only 20 MiB margin and an image-text prefix hit. The immutable v4 artifact publishes the current serving code and version-scoped evidence; historical measurements are not relabeled as runs of that release (Appendix~\ref{app:repro}).

\balance
\section{Conclusion}
JustFit fills a 27B model's native 256K window on a 24 GiB laptop and reaches 320K occupied page-capacity slots across two requests. That high-water is numerically \AggregateRatio{}$\times$ the single-request record of the evaluated historical mlx-vlm configuration, not a like-for-like single-request gain. KVExec bounds reconstruction and checkpointing, PhaseSwap supplies components when needed, and StateTrans preserves history as requests and execution modes change. Complete generation, memory probes, mixed image/text progress, and reasoning evaluation support one design lesson: compact representations become more useful when their execution and storage lifetimes are managed together.

\label{lastmainpage}
\section*{Impact Statement}
User-owned inference can support local control of code and documents and reuse existing hardware. Privacy, energy use, and availability depend on deployment; this study evaluates execution and memory behavior.
\paragraph{Acknowledgments.}
Codex assisted with implementation, tests, and data organization; ChatGPT with writing, LaTeX, and figures. The author is responsible for the code, experiments, and claims.

\clearpage
\nobalance
\begingroup\small
\balance
\setlength{\bibsep}{2pt plus .5pt minus .5pt}\setlength{\parskip}{0pt}
\bibliographystyle{linkedabbrvnat}\bibliography{references}
\endgroup
\clearpage
\appendix
\onecolumn
\renewcommand{\topfraction}{.95}\renewcommand{\textfraction}{.05}
\makeatletter
\renewcommand\normalsize{\@setfontsize\normalsize{10}{11}}
\makeatother
\normalsize
\section{Implementation and Benchmark Details}
\label{app:codec}
\subsection{Codec and Attention Dispatch}
The evaluated geometry uses 256-dimensional heads, four KV heads per full-attention layer, 16 full-attention layers, and 256-token pages. The 48 recurrent layers maintain request-local state. Table~\ref{tab:codec} specifies the TQ4 reconstruction path; the codebook construction and random-sign rule make the representation reproducible without introducing a learned quantizer.

\begin{table}[!htbp]
\centering\small
\caption{TQ4 representation and specialized direct-inverse kernel parameters.}
\label{tab:codec}
\begin{tabular}{L{1.35in}L{4.75in}}
\toprule
Parameter & Value or procedure\\
\midrule
Quantizer & MSE scalar quantization after normalization and randomized Hadamard rotation; K/V use four bits each.\\
Per-vector storage & 32 U32 words (eight indices per word) and one FP16 norm computed from an FP32 norm.\\
Codebook & 16 FP32 centroids for density proportional to $(1-x^2)^{(d-3)/2}$.\\
Construction & 32,768-point grid on $[-1+10^{-6},1-10^{-6}]$; quantile initialization; at most 100 centroid updates; stop when maximum change is below $10^{-6}$.\\
Random signs & NumPy \code{default\_rng(seed + d*7919)}, sampling $\pm1$; K/V default seeds are 0/1.\\
Quantization & Midpoint thresholds between ordered centroids, with strict $>$ comparisons.\\
Inverse kernel & Two radix-16 stages, 16-thread group, 256-element threadgroup float buffer; sequential K and V processing.\\
Launch & Grid $(16,T,g)$ for each group; threadgroup $(16,1,1)$; final path uses $g=2$.\\
Outputs & Contiguous K/V, each $[1,g,T,256]$, in the query dtype; consumed by MLX SDPA for the corresponding query heads.\\
Execution boundary & Call \code{mx.eval} to compute each head-group result before advancing, then the layer result before returning.\\
\bottomrule
\end{tabular}
\end{table}

A \emph{prefill facade} is the request-local cache interface that presents one row of the shared pool to a model call; it is not another KV copy. Single-query attention selects page-native decode. Multi-query attention uses this single-row interface. Direct inverse is selected first when enabled and the mask is a causal string or has compatible array shape. Otherwise, an enabled packed verifier accepts causal-string masks and query lengths 2--4. The remaining fallback gathers the logical packed row for helper prefill, or dequantizes for SDPA. Array-mask eligibility is a shape check. The specialized Metal kernel requires the stated head and page dimensions.

\begin{figure}[htbp]
\centering\resizebox{\textwidth}{!}{%
\begin{tikzpicture}[x=1cm,y=1cm,>=Latex,font=\sffamily\fontsize{9}{11}\selectfont,
 b/.style={draw=plotblue,fill=plotblue!5,align=center,inner sep=5pt,minimum height=1cm},
 p/.style={draw=plotteal,fill=plotteal!5,align=center,inner sep=5pt,minimum height=1cm}]
\node[anchor=west,font=\bfseries] at (0,3) {(a) Attention: reconstruct only the current head group};
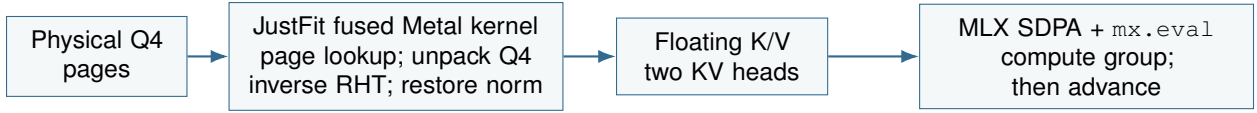
\node[b,text width=2.0cm] (a) at (1.1,1.85) {Physical Q4\\pages};
\node[b,text width=4.0cm] (b) at (5,1.85) {JustFit fused Metal kernel\\page lookup; unpack Q4\\inverse RHT; restore norm};
\node[b,text width=2.4cm] (c) at (9.25,1.85) {Floating K/V\\two KV heads};
\node[b,text width=4.0cm] (d) at (14,1.85) {MLX SDPA + \texttt{mx.eval}\\compute group; then advance};
\draw[->,plotblue,line width=.8pt] (a)--(b);\draw[->,plotblue,line width=.8pt] (b)--(c);\draw[->,plotblue,line width=.8pt] (c)--(d);
\node[anchor=west,font=\bfseries] at (0,.5) {(b) Persistence: borrow storage; retain its owner};
\node[p,text width=4.1cm] (p1) at (2.2,-.65) {Borrow page-run views\\request retains ownership};
\node[p,text width=4.4cm] (p2) at (8.0,-.65) {Synchronous bounded serialization\\single-pass atomic checkpoint};
\node[p,text width=4.1cm] (p3) at (13.8,-.65) {Direct packed restoration\\into the existing page pool};
\draw[->,plotteal,line width=.8pt] (p1)--(p2);\draw[->,plotteal,line width=.8pt] (p2)--node[above,font=\scriptsize,align=center]{later\\request}(p3);
\end{tikzpicture}%
}
\caption{\textbf{KVExec consumer paths.} Grouped prefill reads physical pages and reconstructs only the current two-head group. Prefix persistence borrows page views until synchronous serialization completes; restoration writes packed runs into the existing pool.}
\label{fig:kv-detail}
\end{figure}
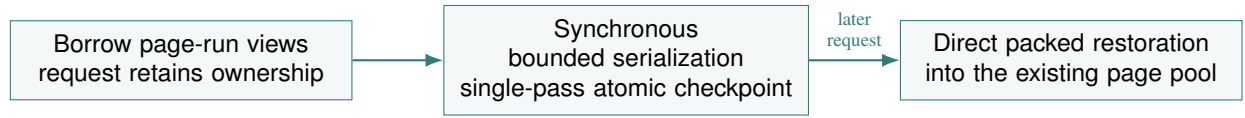
\FloatBarrier
\subsection{Component and Cache Storage}
The evaluated target's non-vision tensors occupy 14,292,384,768 bytes (13.31 GiB), including quantization metadata. The separate BF16 vision encoder/projector occupies 921,460,192 bytes (878.773 MiB). These are tensor payloads, not process-footprint measurements. The embedding and untied head each have U32 packed weights of shape $[248320,640]$ and U8 scales of shape $[248320,160]$, totaling 644.140625 MiB per matrix. The MTP predictor has 215.206 MiB of separate checkpoint payload and references the target vocabulary matrices. Ordinary text execution excludes the vision tower; the image-admission extension temporarily detaches the input embedding and head before image encoding and restores them after image features have been computed. The backbone and packed pool remain attached. These checkpoint sizes describe component payload, while the process measurements include execution and allocation effects.

For each layer and token position, the four heads of either K or V store 512 bytes of packed indices and 8 bytes of norms. Across K/V and 16 attention layers, the result is 16,640 bytes per retained position. The current 262,144-position B1 pool therefore has 4,160 MiB of backing arrays; the 327,680-position B2 pool has 5,200 MiB. The earlier 229,376-position pool used 3,640 MiB. Free-list growth returns page IDs for reuse within this allocation. Recurrent state is separate; the configured BF16 recurrent arrays imply approximately 74.8 MiB per request row before other metadata.

\subsection{Controlled Capacity Protocol}
\label{app:protocol}
The capacity request repeats the word \code{measurement} and asks the model to keep emitting it until the output limit. The checkpoint-local tokenizer applies the chat template, adjusts repetitions to the target length, and checks the actual server prompt count. Temperature is zero, streaming is enabled, and the capacity-only stopping control suppresses EOS. Token IDs 248044 and 248046 receive a logit bias of $-10{,}000$ in the inspected requests. In the updated B1 and evolution cohorts this static bias is applied to both draft and target distributions; an earlier control path did not align them. This controls execution length without replaying a fixed output sequence. A one-token preflight precedes the timed main request.

Earlier cold capacity requests use a fresh isolated prefix-cache directory and no main-request prefix hit. Initialization routes for the current 256K B1 and 320K B2 executions are specified in Appendix~\ref{app:evo10}. All of them perform complete generation after preparation. The earlier 192K+16K B1 cohort uses one lane and a 229,376-position pool; the earlier simultaneous and staggered B2 tests use two lanes, with the latter starting the peer after the first stream's initial token. The controlled 32K+6K study uses the same corpus helper and EOS suppression across configurations. This repeated workload is useful for memory stress and execution comparisons; acceptance rates and throughput are workload-dependent.

\FloatBarrier
\Needspace{110pt}
\section{Additional Evaluation Results}
\label{app:results}
\subsection{Development Configurations}
Table~\ref{tab:evo} identifies the changes behind Figure~\ref{fig:evolution}. Capacity entries record the available completed workload, not a maximum remeasured at every stage. Configuration 8 directly completes 208K B1; the 224K historical B2 frontier remains carried. Configuration 9 has no isolated capacity sweep. Configuration 10 completes 256K B1 and reaches 320K page occupancy in B2; its cold B2 also completes 288K aggregate work. These endpoints have different request shapes. The fixed 32K+6K measurements are a separate series with three processes per stage; APC is disabled for 1--7 and enabled with zero hits for 8--10.

\begin{table}[!htbp]
\centering\small
\caption{Development configurations and their separate fixed-workload results. Capacity entries are in K positions ($K=1{,}024$); the final peak, PP, and TG columns are independent metric medians at 32K+6K. The baseline cross denotes guard termination.}
\label{tab:evo}
\setlength{\tabcolsep}{4pt}
\begin{tabular}{cllrrr}
\toprule
EVO & Added mechanism & Capacity evidence & Peak (MiB) & PP & TG\\
 & & & & \multicolumn{2}{c}{(tok/s)}\\
\midrule
\inputrows data/generated/evolution_rows.tex 
\bottomrule
\end{tabular}
\end{table}

\begin{table}[!htbp]
\centering\small
\caption{Observed min--max ranges for the fixed-workload repeats. Metric medians need not come from the same run. Rounding can make a nonzero range appear constant.}
\label{tab:dispersion}
\begin{tabular}{crrr}
\toprule
EVO & Peak (MiB) & PP (tok/s) & TG (tok/s)\\
\midrule
\inputrows data/generated/dispersion_rows.tex 
\bottomrule
\end{tabular}
\end{table}

\FloatBarrier
\Needspace{110pt}
\subsection{Earlier Fully Cold Single-Request Repeats}
\label{app:limitrepeats}
This earlier configuration-8 cohort, retained as a historical reference, completes 196,608 uncached input positions in each fresh process and 16,384 generated tokens, using one lane. All three output-token sequences agree, and all postflight/page-reuse checks pass. The high-water mark is 13,312 layer-pages: at 16 attention layers and 256 positions per page, this corresponds to 212,992 occupied page-capacity slots. The configured capacity of 14,336 layer-pages corresponds to 229,376 pool slots. Table~\ref{tab:limit-repeats} reports this fresh-process cohort. The prior single observation is retained separately in the artifact archive.

\begin{table}[!htbp]
\centering\small
\caption{Earlier 192K+16K cold repeats, not the final runtime's capacity limit. PP uses uncached tokens over prefill work time; TG uses 16,383 post-first-token outputs. Wall TPS includes all 16,384 outputs and the full reported request interval. Margin is the sampled distance below the 21,000-MiB guard.}
\label{tab:limit-repeats}
\begin{tabular}{crrrrrr}
\toprule
Run & PP (tok/s) & TG (tok/s) & Wall (s) & Wall TPS & Peak (MiB) & Margin (MiB)\\
\midrule
1 & 68.321 & 7.0149 & 5,213.943 & 3.1423 & 20,862 & 138\\
2 & 68.264 & 7.0158 & 5,216.103 & 3.1410 & 20,857 & 143\\
3 & 68.287 & 7.0177 & 5,214.499 & 3.1420 & 20,959 & 41\\
\bottomrule
\end{tabular}
\end{table}

\subsection{Earlier Concurrent Workloads}
\begin{table}[!htbp]
\centering\small
\caption{Earlier concurrent operating points. The current 288K/320K B2 results appear in Appendix~\ref{app:evo10}. Each row completes all outputs and passes postflight/reuse. TG is common-interval aggregate. Earlier B4 wall rates are omitted because their singleton-tail token-control protocol differs from the current run.}
\label{tab:concurrent}
\begin{tabular}{lrrrr}
\toprule
Workload / arrival & Positions & PP & TG & Peak (MiB)\\
\midrule
B2 $2\times$(96K+12K), simultaneous & 221,184 & 90.51 & 12.72 & 20,195\\
B2 $2\times$(96K+12K), staggered & 221,184 & 90.52 & 12.81 & 20,156\\
B2 $2\times$(96K+16K), simultaneous & 229,376 & 90.44 & 12.55 & 20,310\\
B2 $2\times$(96K+16K), staggered & 229,376 & 90.44 & 12.51 & 20,247\\
B4 96K+12K; $3\times$(8K+4K) & 147,456 & 91.25 & 23.41 & 19,178\\
B4 128K+12K; $3\times$(8K+4K) & 180,224 & 83.35 & 21.54 & 20,236\\
B4 128K+12K; $3\times$(8K+12K) & 204,800 & 83.16 & 20.78 & 20,623\\
\bottomrule
\end{tabular}
\end{table}

\FloatBarrier
\Needspace{130pt}
\subsection{Current Four-Stream Qualification}
\label{app:b4current}
The released scheduler removes the default LM-head suffix gate while retaining lane and paged-KV reservations, the 8K output guarantee, the 16K output ceiling, and N4/PF64 interleaving. This fresh-process, zero-prefix-hit trial cold-builds a 131,072-token prompt, then submits three 8,192-token peers after its first output. All four streams complete 12,288 outputs. EOS bias is aligned between draft and target, the head residency policy is enabled, and MTP is used only for an eligible singleton. No head-phase deferrals or guard trips occur. The test exercises a long incumbent, not four simultaneously submitted long prompts.

\begin{table}[!htbp]\centering\small
\caption{Current staggered B4, one completed trial. Common-active TG and full-request wall rate have different denominators.}
\label{tab:b4current}
\begin{tabular}{lr}\toprule
Quantity & Observation\\\midrule
Total uncached input / completed output tokens & 155,648 / 49,152\\
Prefill work / PP & 1,821.472 s / 85.452 tok/s\\
Common-active outputs / interval & 42,945 / 1,902.766 s\\
Common-active aggregate TG & 22.570 tok/s\\
First-request start to last completion / wall rate & 4,046.521 s / 12.147 tok/s\\
Sampled peak / guard margin & 18,915 / 2,085 MiB\\
Occupied high-water / configured layer-pages & 12,608 / 12,800\\
Postflight / page reuse & pass / pass\\\bottomrule
\end{tabular}\end{table}

The 204,800 input-plus-output positions are cumulative completed work, not simultaneous occupancy. The page high-water corresponds to 201,728 position slots, at 256-token page granularity, in a 204,800-position pool. All output-token sequences agree. This qualification establishes completion and four-stream overlap under the stated protocol; different arrivals and token controls prevent attributing its difference from earlier B4 rows solely to the gate change.

\FloatBarrier
\subsection{Earlier Short-Output Probes}
\label{app:short}
The short-output cohort generates 64 tokens per request. Its common 8K input point gives a direct baseline/JustFit comparison; the other completing lengths support the process-growth fit in Figure~\ref{fig:hero}. The pool is sized per probe. The fitted intercept absorbs the process offset, while the slope includes allocation and execution effects.
\begin{table}[!htbp]\centering\small
\caption{Earlier B1 short-output probes. TG excludes the first generated token. A cross indicates guard termination.}
\label{tab:short}
\begin{tabular}{llrrr}\toprule
Input & Runtime & Peak (MiB) & PP (tok/s) & TG (tok/s)\\\midrule
8K & mlx-vlm & 17,180 & 129.65 & 15.63\\
8K & JustFit & 15,746 & 127.27 & 24.31\\
16K & mlx-vlm & 18,449 & 125.23 & 14.97\\
24K & mlx-vlm & 19,898 & 121.12 & 14.97\\
32K & mlx-vlm & $\times$ & --- & ---\\
32K & JustFit & 15,899 & 114.76 & 19.11\\
64K & mlx-vlm & $\times$ & --- & ---\\
64K & JustFit & 16,433 & 101.42 & 14.59\\\bottomrule
\end{tabular}\end{table}
\FloatBarrier
\subsection{Residency-Manager Refactor}
\label{app:regression}
Table~\ref{tab:regression} compares one final pair per workload before and after the common-manager refactor. The staggered B2 test exercises MTP-to-AR-to-MTP transitions. All per-stream outputs agree, and postflight/page-reuse checks pass. Both arms use PhaseSwap. The B4 peak difference is reported alongside the earlier 19,159-MiB pre-refactor observation to show run-to-run variation.

\begin{table}[!htbp]
\centering\small
\caption{Short paired regression. Each arrow is component-specific manager $\rightarrow$ common residency manager.}
\label{tab:regression}
\begin{tabular}{lrrr}
\toprule
Workload & PP (tok/s) & TG (tok/s) & Peak (MiB)\\
\midrule
B1 24K+512 & $118.757\rightarrow118.410$ & $19.246\rightarrow19.243$ & $18{,}832\rightarrow18{,}830$\\
B4 $4\times$(2K+512) & $121.538\rightarrow122.960$ & $37.762\rightarrow39.305$ & $19{,}382\rightarrow19{,}157$\\
B2 2K each, 512/128 output & $125.822\rightarrow125.399$ & $30.367\rightarrow30.110$ & $19{,}001\rightarrow19{,}003$\\
\bottomrule
\end{tabular}
\end{table}

\FloatBarrier
\section{Reasoning Evaluation and Reproducibility}
\label{app:quality}
\subsection{AIME Protocol and Error Cases}
The paired MacBook evaluation uses an EvalScope-style prompt requesting step-by-step reasoning and repeating the instruction to place the final answer in \code{\textbackslash boxed\{\}}. Sampling uses temperature 1, top-$p$ 0.95, top-$k$ 20, and seed 20260811, with the runtime's \code{xhigh} reasoning setting. Both arms enable MTP with draft block size 3 and disable prefix caching. Each arm reports one seeded evaluation over the same 30 problems.

Output ceilings are 71,680 tokens for TQ4 Q1--Q14 and 98,304 for TQ4 Q15--Q30; INT8 uses 71,680 throughout. TQ4 uses the page-backed storage and attention path, whereas INT8 uses the existing non-paged reference path. Together with the different ceilings and recovery procedure below, this makes the comparison an integrated capability check rather than an isolated KV-bit-width experiment.

Paged TQ4 completes the 30-problem AIME 2026 evaluation with 29 correct answers. The only failure reaches its 98,304-token output ceiling. Uniform INT8 answers 28/30; after a 21,153-MiB guard event following Q10, Q11--Q30 run on fresh servers. Both arms reach an output ceiling on Q15. On Q19, INT8 returns 271, omitting the one-digit cases 1--8, whereas TQ4 returns 279. This is a trajectory-level difference, not evidence that TQ4 is generally more accurate.

The TQ4 suite produces 696,834 output tokens at 15.04 token-weighted tokens/s; the arithmetic mean of per-question decode rates is 17.59. INT8 produces 680,071 at 14.64 token-weighted tokens/s, with a mean per-question rate of 17.84. The different denominators should not be interchanged. The separate MacBook Air/Bonsai run answers 26/30, producing 1,057,606 output tokens at 5.22 token-weighted tokens/s. Host suspension between its records excludes suite wall time from the comparison. These capability runs are distinct from the repetitive capacity tests and from the later single-request repetitions.

\subsection{Reproducibility Materials}
\label{app:repro}
The \href{https://github.com/YuhuaBillChen/justfit-mlx/tree/justfit-repro-v4/examples/justfit}{immutable reproduction tag \code{justfit-repro-v4}} provides the current serving code, component-download instructions, launcher, benchmark client, and public result files. Its \href{https://github.com/YuhuaBillChen/justfit-mlx/blob/justfit-repro-v4/examples/justfit/quickstart.md}{Quick Start} covers direct API use, Open WebUI, and a coding-agent configuration. The release includes EVO10 persistence, image/input-embedding residency, and removal of the default mixed-prefill suffix gate. The current four-stream qualification is published as \code{b4-throughput.json}; its runtime identity and timing definitions accompany the measurements. A passing qualification on the tested host is not a claim that every installation or workload has been validated.

A supplementary public-kit reconstruction on an M3 Pro (36 GiB) completed three fresh-process 32K+6K repeats and one run each of cold 240K+16K B1, cold $2\times$(128K+16K) B2, and the four-stream incumbent workload under the same 21,000-MiB process-footprint guard; protocols, results, and host/scheduling differences are documented in the companion repository's \href{https://github.com/YuhuaBillChen/justfit-mlx/tree/main/examples/justfit/results/m3-pro}{supplementary device results}, and equal process guards do not imply equal host-memory conditions.

Table~\ref{tab:m3-m4} puts the completed M3 runs beside the published M4 results. The fixed 32K+6K row uses independent three-process metric medians on each host; each long M3 workload has one completed run. PP divides uncached input by summed prefill work time. B1 TG excludes the first output token; B2 and B4 TG are aggregate common-active rates, not per-request rates. Peak is sampled physical process footprint. This is not a controlled chip-only comparison: the M3 has 36 GiB versus 24 GiB on the M4, and host pressure, preparation, and B2 arrival/occupancy differ. Equal 21,000-MiB process guards do not establish equal remaining system memory.

\begin{table}[!htbp]
\centering\small
\caption{Supplementary public-kit M3 Pro (36 GiB) results beside published M4 Pro (24 GiB) results. B4 is one 128K incumbent followed by three 8K peers, each generating 12K tokens. Fixed-workload entries are independent medians ($n=3$ per host); the long M3 entries are single observations. The comparison is descriptive, not an isolated hardware speedup.}
\label{tab:m3-m4}
\setlength{\tabcolsep}{4pt}
\begin{tabular}{lrrrrrr}
\toprule
 & \multicolumn{2}{c}{PP (tok/s)} & \multicolumn{2}{c}{TG (tok/s)} & \multicolumn{2}{c}{Peak (MiB)}\\
Workload & M3 & M4 & M3 & M4 & M3 & M4\\
\midrule
\inputrows data/generated/m3_comparison_rows.tex
\bottomrule
\end{tabular}
\end{table}
\FloatBarrier

A tensor-level audit matches all 1,349 non-vision tensors of the local target to the pinned public MXFP4 checkpoint, all 333 production BF16 vision tensors to the published component, and the MTP weights by digest. The container differs in inactive checkpoint-side vision storage.

The manuscript package includes editable figures and compilation data. The public result archive preserves the 256K/320K evidence separately from the earlier 192K+16K cohort. Release code and historical experiment provenance are distinct: the new tag makes the kit runnable but does not retroactively change which runtime produced an older result. The artifact manifests retain source identities, component digests, protocols, and available environment metadata for each cohort.

\FloatBarrier
\Needspace{110pt}
\section{Decode-Mode Selection: AR versus Batched MTP}
\label{app:batchedmtp}
The serving policy retains singleton MTP and uses ordinary AR batching when peers are active. To test this choice, we compare AR without a drafter against an experimental batched-MTP path at B2, B3, and B4. Each row has either 2,048 input and 512 output tokens, or 32,768 input and 2,048 output tokens. Every arm uses a fresh process, greedy selection, matched pool capacity within its pair, and EOS suppression aligned between draft and target. APC is enabled with a fresh namespace and zero hits, not disabled. The cadence is N1/PF256 (one decode step and prefill chunks up to 256 tokens), not the mixed-serving N4/PF64 profile. Short and long suites use different backing volumes; only within-pair comparisons are controlled.

The experimental patch separates drafter input padding from target-history padding and clears allocator cache at a safe cohort-shrink boundary. It passes 978 focused tests with two pre-existing strict expected failures. All 12 measured arms complete without a guard trip and pass postflight/reuse; output-token hashes match in each AR/MTP pair. There is one fresh-process observation per arm, so small differences, especially at B2, do not establish statistical significance. These synthetic workloads are not application-level latency or quality evaluations.

\begin{figure}[htbp]
\centering\begin{minipage}{.49\linewidth}\centering
\begin{tikzpicture}
\begin{axis}[paperaxis,width=\linewidth,height=2.0in,xmin=1.6,xmax=4.4,ymin=-25,ymax=5,
xtick={2,3,4},xlabel={Concurrent requests},ylabel={MTP vs. AR throughput (\%)},
legend style={at={(.5,1.03)},anchor=south,legend columns=2}]
\addplot[only marks,plotblue,mark=*,mark size=3pt] coordinates {(2,-1.735) (3,-18.375) (4,-20.407)};
\addplot[only marks,plotorange,mark=square*,mark size=3pt] coordinates {(2,-1.903) (3,-12.161) (4,-12.066)};
\legend{2K+512,32K+2K}
\addplot[gray,dashed,forget plot] coordinates {(1.6,0) (4.4,0)};
\node[font=\scriptsize,anchor=south east] at (axis cs:4.35,0) {AR parity};
\end{axis}\end{tikzpicture}
\end{minipage}\hfill
\begin{minipage}{.49\linewidth}\centering
\begin{tikzpicture}
\begin{axis}[paperaxis,width=\linewidth,height=2.0in,xmin=1.6,xmax=4.4,ymin=0,ymax=1800,
xtick={2,3,4},xlabel={Concurrent requests},ylabel={Additional MTP peak (MiB)},
legend style={at={(.5,1.03)},anchor=south,legend columns=2}]
\addplot[only marks,plotblue,mark=*,mark size=3pt] coordinates {(2,597) (3,936) (4,1228)};
\addplot[only marks,plotorange,mark=square*,mark size=3pt] coordinates {(2,596) (3,1071) (4,1521)};
\legend{2K+512,32K+2K}
\end{axis}\end{tikzpicture}
\end{minipage}
\caption{\textbf{Why the current policy keeps concurrent decode on AR.} Each point is one matched pair, not an average over repetitions. Left: aggregate common-interval TG change relative to AR. Right: difference in sampled whole-process peaks, not MTP weight size. B2 is near parity; B3/B4 are slower and require more peak memory in this implementation.}
\label{fig:batchedmtp}
\end{figure}
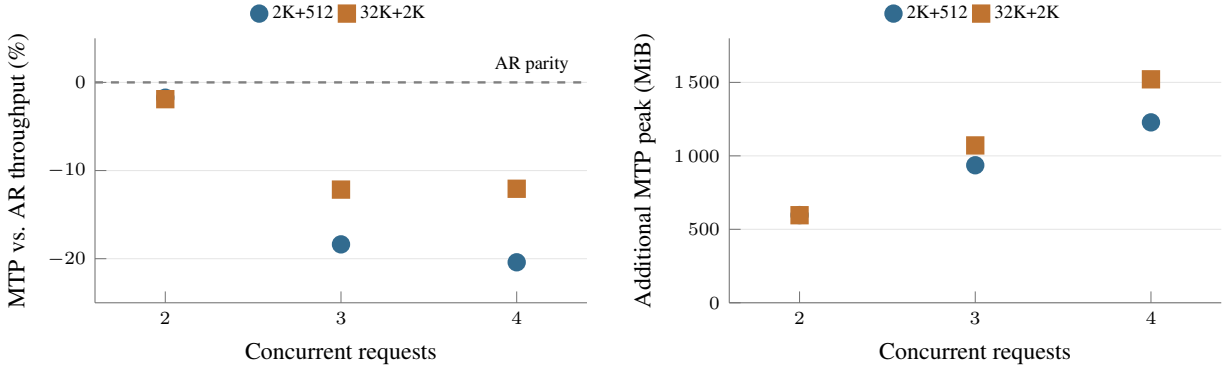

\begin{table}[htbp]\centering\small
\caption{Batched-MTP experiment. TG is aggregate over the common active interval, not per-request or full-request wall throughput. Input/output lengths are per request.}
\label{tab:batchedmtp}
\begin{tabular}{lrrrrr}\toprule
Profile & AR TG & MTP TG & Change (\%) & AR peak (MiB) & MTP peak (MiB)\\\midrule
B2, 2K+512 & 30.455 & 29.926 & $-1.74$ & 14,946 & 15,543\\
B3, 2K+512 & 38.170 & 31.156 & $-18.38$ & 15,276 & 16,212\\
B4, 2K+512 & 39.980 & 31.822 & $-20.41$ & 16,041 & 17,269\\
B2, 32K+2K & 21.322 & 20.916 & $-1.90$ & 16,505 & 17,101\\
B3, 32K+2K & 25.405 & 22.315 & $-12.16$ & 17,284 & 18,355\\
B4, 32K+2K & 26.138 & 22.984 & $-12.07$ & 18,017 & 19,538\\\bottomrule
\end{tabular}\end{table}

\paragraph{Interpretation and remaining work.}
Acceptance samples at the actual concurrent batch sizes reach 100\% for this repetitive workload; high acceptance alone does not guarantee a throughput gain. Verification width $S=3$ means two drafted positions plus a bonus position. Source inspection shows one target call on a $[B,S]$ input, not $S$ complete target calls. An isolated quantized-linear sweep confirms some amortization across $S$, while exposing poorer scaling with $B$. Thus the measurements do not establish serialized whole-model verification or a correctness defect specific to MXFP4.

The microbenchmark covers three model-sized projection geometries, B1--B4, widths 1--4, and MXFP4/group-32 versus affine Q4/group-32 and group-64: 144 cases, each checked against its own format's singleton reference. For the $5120\rightarrow17408$ projection at B2/$S=3$, median times are 0.704 ms for MXFP4 and 0.643 ms for affine/group-64; native AR is 0.353 ms for MXFP4. The format comparison uses independently quantized versions of the same synthetic dense matrix, not full-model checkpoints or cross-format output parity. It is diagnostic evidence, not an affine-Q4 end-to-end speed claim.

Further work should isolate verifier projection dispatch, batch tiling, recurrent-state preparation/commit, and head evaluation under matched cache state. The current evidence supports a conservative serving policy for this runtime and hardware; it does not imply that an optimized batched-MTP implementation cannot outperform AR. During the long suite, host-wide telemetry records no additional swap-out, but pre-existing swap and swap-in remain; we do not describe these runs as having zero swap.

\FloatBarrier
\Needspace{110pt}
\section{Current Full-Window and Shared-Capacity Experiments}
\label{app:evo10}
The final configuration is evaluated at 262,144 positions for one request and 327,680 occupied page-capacity slots across two requests. All completed executions use a fresh process, repetitive text, greedy selection, aligned EOS suppression, and a 21,000-MiB process guard. The reported capacity is reached while performing complete generation, without runtime truncation, sliding-window eviction, or summarization. Recurrent model state remains separate from the full-attention KV pool.

\subsection{Three completed B1 native-window executions}
All three executions complete the same 245,760-input/16,384-output shape, generate identical output tokens, and pass postflight and page reuse. Their 16,384 occupied layer-pages correspond to the complete 262,144-position pool. Table~\ref{tab:native-window} groups them by that shared endpoint. The first constructs the whole input cold; reruns initialize from a saved 161,792-position prefix and compute another 83,968 positions before repeating the entire 16K generation. Each prefix-initialized trial reruns the complete generation interval.

\begin{table}[htbp]\centering\small
\caption{Current B1 240K+16K executions. PP measures the newly computed input in its own row; TG always measures 16,383 post-first-token outputs at the same final context shape. The checkpoint rows are complete generation reruns.}
\label{tab:native-window}
\setlength{\tabcolsep}{4pt}
\begin{tabular}{clrrrrrr}\toprule
Run & Preparation & Cached & New input & PP & TG & Peak (MiB) & Margin\\
 & & & & \multicolumn{2}{c}{(tok/s)} & & (MiB)\\\midrule
\inputrows data/generated/native_rows.tex 
\bottomrule
\end{tabular}\end{table}
Median TG across the three complete executions is \NativeTG{} tokens/s; the range is 5.936--5.987. Median sampled peak is 20,449 MiB, with a 20,357--20,450-MiB range and at least 550 MiB of sampled margin. Whole-request intervals are 6,820.029, 4,632.095, and 4,633.636 seconds, reflecting the different input preparation and therefore not pooled as cold latency.

\paragraph{Shared-prefix timing reconstruction.}
The first cold trace reaches 161,792 positions after approximately 2,163.516 seconds. Combining that measured prefix interval with each rerun's incremental interval gives
\begin{equation}
 \widehat{\mathrm{PP}}_{240K,r}=\frac{245{,}760}{T_{\mathrm{prefix},1}+T_{\mathrm{suffix},r}}.
\label{eq:ppreconstruct}
\end{equation}
The suffix intervals, 1,894.100 and 1,894.797 seconds, yield 60.568 and 60.557 tokens/s, compared with measured cold PP of 60.546. The reconstructed rates agree with the directly measured cold rate within 0.04\%. This calculation shares the measured prefix interval and excludes restoration, providing a timing-consistency check.

\paragraph{Host persistence intervals.}
The two checkpoint-initialized B1 executions record exact-cache-file load calls of 0.0464 seconds and sums of 16 packed page-run read calls of 0.5332 and 0.5380 seconds. Outer checkpoint writes take 0.8288/0.9644 and 0.8128/0.7803 seconds; cold-run outer writes take 0.859/0.918 seconds. Row-normalization calls are individually below 1 ms. These are host-call observations, not a complete end-to-end restore decomposition or measurements of completed GPU work and physical SSD durability.

\subsection{Three completed B2 320K aggregate executions}
All three executions complete (160K+16K)+(128K+16K), reach 20,480 occupied layer-pages (327,680 page-capacity slots), match per-lane output tokens, and pass postflight/reuse. Each common active interval covers 32,763 generated tokens. Decode intervals are 3,366.468, 3,364.545, and 3,365.639 seconds; the corresponding aggregate common-active-interval TG median is \AggregateTG{} tokens/s. Table~\ref{tab:warm-b2} reports generation throughput alongside the preparation state for each complete execution.

\begin{table}[htbp]\centering\small
\caption{Current B2 320K aggregate executions. Every lane generates 16,384 outputs. TG is common-active-interval aggregate throughput. Prefix coverage changes input work and whole-run peak, not the final workload shape.}
\label{tab:warm-b2}
\setlength{\tabcolsep}{5pt}
\begin{tabular}{crrrrr}\toprule
Run & Cached A / B & New A / B & TG (tok/s) & Peak (MiB) & Margin (MiB)\\\midrule
\inputrows data/generated/b2_rows.tex 
\bottomrule
\end{tabular}\end{table}
The first execution computes 10,241 input positions in 197.952 seconds; reruns compute two positions in 0.492 and 0.423 seconds. Those short residual intervals are not long-prefill speed measurements. The first attempt's peak exceeds the reruns by 497--521 MiB, so these peaks remain associated with their preparation routes. Whole-request intervals are 3,567.037, 3,367.731, and 3,368.731 seconds.

The longer lane is submitted first and the second 0.5 seconds later. Each rerun uses an independent clone of the frozen prefix base. This ordered admission limits overlapping preparation. The next tested shape, (168K+16K)+(128K+16K), requests 335,872 positions and trips the guard at 21,097 MiB during restoration. Operator-cancelled probes are excluded from the measured capacity outcomes.

\subsection{Fully cold B2 construction}
A separate fresh process completes two 131,072-input/16,384-output requests, totaling 294,912 positions. It processes all 262,144 input tokens at 81.399 tokens/s, then attains 10.341 common-interval aggregate TG and 5.113 wall-output tokens/s. Peak is 20,858 MiB. The instantaneous high-water is 18,400 layer-pages, equivalent to 294,400 positions, below the 294,912-position pool because the requests finish at different times. These measurements retain the cold-construction path as an additional completed workload; the headline aggregate record is the 320K B2 shape.

\FloatBarrier
\Needspace{130pt}
\subsection{Controlled mechanism probes}
\label{app:mechanisms}
Head grouping is measured on synthetic single-layer attention without model weights: BF16 queries, Q4 K/V, 24 query heads, four KV heads, dimension 256, 256-position pages, and query chunk 256. The protocol uses reversed physical-page mappings, a poison page, varying norms, and tail-page correctness checks. Each path has six timing samples with alternating order and cache clearing. The tested two-head grouping matches the all-head reference bitwise at the reported target lengths. An optional one-head diagnostic differed at a tail length and was not used for the selected two-head implementation.

\begin{table}[htbp]\centering\small
\caption{All-head versus two-head-group prefill. Latency includes reconstruction, group evaluation/synchronization, SDPA, and concatenation. Peak increments include SDPA workspace, not just raw K/V operands.}
\label{tab:headgroups}
\begin{tabular}{lrrrr}\toprule
History & All-head (ms) & Two-head (ms) & All-head peak (MiB) & Two-head peak (MiB)\\\midrule
\inputrows data/generated/head_group_rows.tex 
\bottomrule
\end{tabular}\end{table}
At 192K, incremental MLX peak decreases by approximately 49.1\% while latency increases 3.6\%. Mathematical independence of attention heads permits the split; measured equality and complete-model tests establish the selected implementation's behavior.

Snapshot probes use 16 packed cache layers without model weights. At 32K and 224K prefixes, detached row extraction peaks at 520 and 3,640 MiB incremental MLX memory; synchronous borrowed views reduce this to 32.5 and 227.5 MiB, respectively. The remaining staging is layer-scoped rather than a second full cache. Synthetic initialization can dominate absolute process footprint, so these incremental allocation results are not treated as full-model physical-memory savings.

\begin{table}[htbp]\centering\small
\caption{Writer-only medians from three old/new pairs per size with alternating order. Application bytes count temporary-file and destination writes, not physical NAND traffic.}
\label{tab:writer}
\setlength{\tabcolsep}{5pt}
\begin{tabular}{lrrrr}\toprule
Prefix & Old save (s) & Single-pass (s) & Old application bytes & Single-pass bytes\\\midrule
\inputrows data/generated/writer_rows.tex 
\bottomrule
\end{tabular}\end{table}
Old/new incremental MLX peaks are 16.25/8, 32.5/16, and 227.5/112 MiB for the three writer sizes. The writer probe and the snapshot-borrowing probe isolate different operations. Fixed 32K+6K tests show nearly unchanged sustained decode in the final stages, while full-model capacity tests establish the larger usable envelope of the combined design.

\FloatBarrier
\Needspace{110pt}
\section{Phase-Separated Image Admission}
\label{app:vision}
\subsection{Serving Contract and Admission Policy}
An eligible Qwen image joins at a serialized boundary, preserving incumbent target KV pages and Gated DeltaNet rows. Admission requires age at least 500 ms, a free lane within the four-lane limit, a fitting page reservation, and active retained state within the media gate. The later candidate gate is 198,656 positions (196,608 input plus 2,048 headroom); it counts actual input-plus-generated state, not reserved output. Overdue images block further text bypass that turn. Selected prefill uses N4/PF64; each batched forward advances all active AR rows.

On a feature-cache miss, the path finishes outstanding work, releases head/embedding generation leases, detaches the unowned embedding, and acquires the tower. After feature evaluation it releases the tower and restores embedding/head. The chunked provider combines features with text embeddings, resolving the table through its owner rather than retaining an old table. Feature-cache hits bypass tower loading. Backbone and paged pool remain resident.

\subsection{Workload and Measurements}
Each fresh isolated server uses a cloned frozen APC directory. Repeated-word text uses greedy decoding, aligned stopping-token suppression, and 512 outputs. One $1049\times1176$ PNG arrives about 0.75 s after the first text token, with a 524,288-pixel area limit, 2,048-pixel edge limit, and microbatch one. Its 453-position prompt generates 64 outputs. Both streams must complete, image first. Table~\ref{tab:vision-full} gives the experimental gates: 143,360 for 96K/128K and 204,800 for 160K/192K/196K. At admission the 192K arm fits the later 198,656-position candidate; the 200,704-input edge does not.

\begin{table}[htbp]\centering\small
\caption{\textbf{Complete mixed text/image ladder.} Text input, cached, and new counts are positions. Gate is the experimental active-retained limit; guard is whole-process MiB. Image E2E covers the full request. One observation per row; each image finishes before the text output.}
\label{tab:vision-full}
\setlength{\tabcolsep}{3.2pt}
\begin{tabular}{rrrrrrrr}\toprule
Text input & Cached & New input & Gate & Guard & Peak & Margin & Image E2E\\
 & & & (positions) & (MiB) & (MiB) & (MiB) & (s)\\\midrule
\inputrows data/generated/vision_all_rows.tex
\bottomrule\end{tabular}\end{table}

All five arms load the tower and encode the image; all text outputs match. At 200,704 input, the image's text prefix restores 452/453 positions, contributing to its 12.994-s E2E. This is a boundary trial; an earlier operator-stopped preparation at this length is excluded.

\begin{table}[htbp]\centering\small
\caption{Text-side timing in the same mixed-image trials. TTFT includes prefix restoration and newly computed input; stream gap is the recorded maximum during streaming.}
\label{tab:vision-text}
\begin{tabular}{rrr}\toprule
Text input & TTFT (s) & Maximum stream gap (s)\\\midrule
\inputrows data/generated/vision_timing_rows.tex
\bottomrule\end{tabular}\end{table}

The 196,608-input trial records embedding MLX-active allocation falling from 18,354.28 to 17,710.14 MiB, a 644.14-MiB change. Tower load, embedding restore, and head restore take 0.151, 0.113, and 0.110 s in host intervals. These observations identify the intended residency transition, while the independent whole-process peak is 20,852 MiB.

\subsection{Failure Handling and Validation}
Focused validation passes 475 runtime tests and six launcher-policy tests, covering feature-cache bypass, feature-first execution, weak-reference reclamation, restored-table lookup, and failure cleanup. Other owners prevent unsafe unload. Failed encoding restores text leases; failed tower release blocks further large allocations; a missing embedding artifact stops startup. The final routing change keeps video on the existing path and was unit-tested after the image model runs. Section~\ref{sec:limits} discusses the experimental scope.

\end{document}